\documentclass{ecai} 

\usepackage{hyperref}
\usepackage{latexsym}
\usepackage{amssymb}
\usepackage{amsmath}
\usepackage{amsthm}
\usepackage{booktabs}
\usepackage{enumitem}
\usepackage{graphicx}
\usepackage{color}

\newcommand{\BibTeX}{B\kern-.05em{\sc i\kern-.025em b}\kern-.08em\TeX}

\begin{document}


\begin{frontmatter}


\paperid{5183} 


\title{Beyond the Basics: Leveraging Large Language Model for Fine-Grained Medical Entity Recognition}


\author[A]
{\fnms{Nwe Ni}~\snm{Win}\orcid{0009-0004-0552-8338}\thanks{Corresponding Author. Email: 22132742@student.westernsydney.edu.au}}

\author[A,B]{\fnms{Jim}~\snm{Basilakis}\orcid{0000-0002-7440-1320}}

\author[B]
{\fnms{Steven}~\snm{Thomas}\orcid{0000-0002-2416-0020}}

\author[C,D]
{\fnms{Seyhan}~\snm{Yazar}\orcid{0000-0003-0994-6196}}

\author[D]{\fnms{Laura}~\snm{Pierce}\orcid{0000-0000-0000-0000}}

\author[E]{\fnms{Stephanie}~\snm{Liu}\orcid{0000-0000-0000-0000}}

\author[B]{\fnms{Paul M. }~\snm{Middleton}\orcid{0000-0003-0760-1098}}

\author[B]{\fnms{Nasser}~\snm{Ghadiri}\orcid{0000-0002-6519-6548}}

\author[A,B]
{\fnms{X. Rosalind}~\snm{Wang}\orcid{0000-0001-5454-6197}}

\address[A]{Western Sydney University, Australia}
\address[B]{South Western Emergency Research Institute, Australia}
\address[C]{Garvan Institute of Medical Research, Australia}
\address[D]{University of New South Wales, Australia}
\address[E]{Liverpool Hospital, Australia}


\begin{abstract}
Extracting clinically relevant information from unstructured medical narratives such as admission notes, discharge summaries, and emergency case histories remains a challenge in clinical natural language processing (NLP). Medical Entity Recognition (MER) identifies meaningful concepts embedded in these records. Recent advancements in large language models (LLMs) have shown competitive MER performance; however, evaluations often focus on general entity types, offering limited utility for real-world clinical needs requiring finer-grained extraction.
To address this gap, we rigorously evaluated the open-source LLaMA3 model for fine-grained medical entity recognition across 18 clinically detailed categories. To optimize performance, we employed three learning paradigms: zero-shot, few-shot, and fine-tuning with Low-Rank Adaptation (LoRA). To further enhance few-shot learning, we introduced two example selection methods based on token- and sentence-level embedding similarity, utilizing a pre-trained BioBERT model.
Unlike prior work assessing zero-shot and few-shot performance on proprietary models (e.g., GPT-4) or fine-tuning different architectures, we ensured methodological consistency by applying all strategies to a unified LLaMA3 backbone, enabling fair comparison across learning settings.
Our results showed that fine-tuned LLaMA3 surpasses zero-shot and few-shot approaches by 63.11\% and 35.63\%, respectivel respectively, achieving an F1 score of 81.24\% in granular medical entity extraction.

\end{abstract}

\end{frontmatter}


\section{Introduction}

\begin{figure}[tb]
\includegraphics[width=\linewidth]{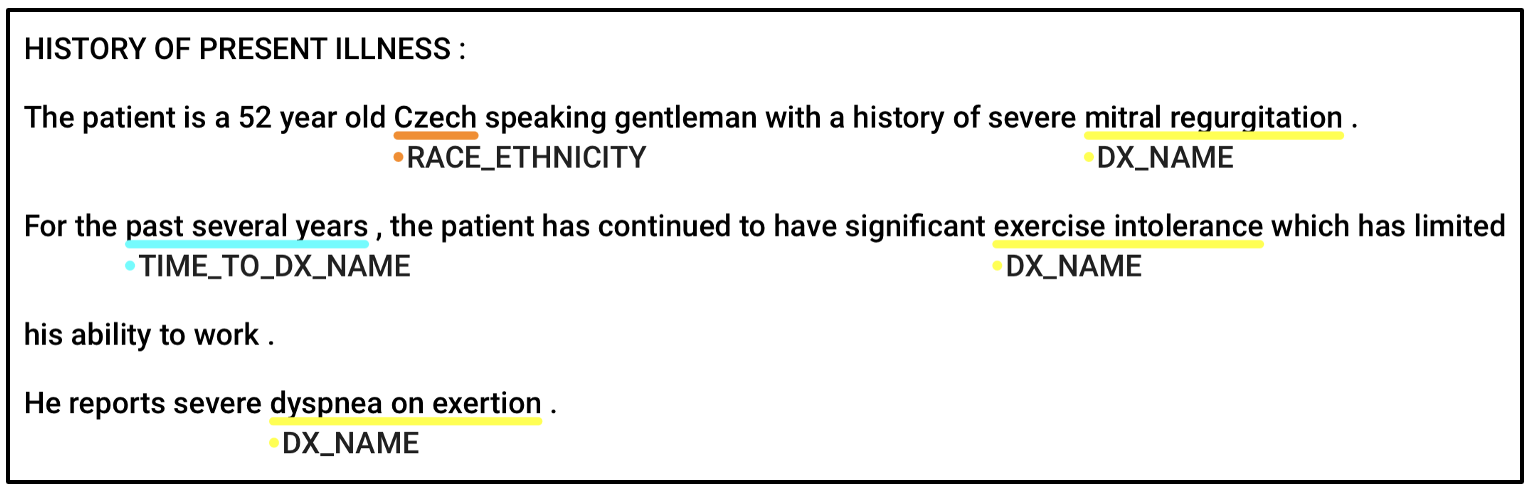}
\caption{Example of Medical Entity Recognition from Unstructured Medical Text}
\label{fig:mer_eg}
\vspace{8mm}
\end{figure}

Electronic Medical Record (EMR) systems store patient information and medical conditions as digital records. The volume of patient data collected by hospitals and other medical institutions is substantial and continues to grow. It is currently estimated that approximately 75\% of this data is unstructured~\cite{capurro_availability_2014}, making it challenging to analyse using traditional data processing tools. As a result, the rich clinical information embedded in EMRs remains largely underutilised. Extracting structured information from unstructured medical text is a foundational task in clinical NLP~\cite{jensen_mining_2012,raza_large-scale_2022}. This process, often referred to as medical entity recognition (MER) or named entity recognition (NER), involves extracting and classifying key medical entities such as medications, diagnoses, and procedures embedded within free-text narratives~\cite{nadkarni_natural_2011} as illustrated in Figure~\ref{fig:mer_eg}. Accurate MER is critical for a range of downstream applications, including clinical decision support for better patient care, clinical research, patient cohort identification, medical question-answering systems, and constructing longitudinal patient medical histories~\cite{jensen_mining_2012,zhou_clinical_2020,sun_evaluating_2013}.

Traditional MER apporaches have relied heavily on hand-crafted features and domain-specific ontologies, such as the Unified Medical Language System (UMLS)~\cite{bodenreider_unified_2004}. In recent years, deep learning methods, and more recently LLMs, have significantly advanced the state of the art in NER tasks in different domains by learning contextual representations directly from the data. General-purpose LLMs such as GPT3.5 and GPT-4~\cite{openai_gpt-4_2024}, have shown promise for zero-shot and few-shot NER tasks~\cite{hu_improving_2024}. In parallel, research has explored few-shot prompting techniques, using example selection methods to enhance NER performance~\cite{wang_gpt-ner_2023}, and performance comparisons have been made against supervised fine-tuned models on architectures such as BERT~\cite{devlin_bert_2019}. 

Despite these advancements, important gaps still exist. Most studies focus on high-level medical entities like Problem, Test, or Treatment using benchmark datasets which do not capture the granularity required in real clinical applications. Furthermore, past researchers have used open-source LLMs, however, their evaluation setups are often inconsistent, comparing few-shot performance from proprietary models like GPT-4 with fine-tuned results from open-source models. This makes it hard to fairly compare the relative effectiveness of different learning paradigms.
To the best of our knowledge, this is the first study to directly address these gaps. We provide a thorough evaluation of fine-grained MER using the open-source LLaMA3 8B Instruct model as a single backbone across three learning settings: zero-shot, few-shot, and fine-tuning. This consistent setup allows for fair and reliable comparisons. The contributions of our work are summarized as follows:

\begin{itemize}
\item We developed a fine-grained MER dataset based on discharge summaries from the i2b2\footnote{\url{https://www.i2b2.org/NLP/DataSets/}} corpus, annotated across 18 clinically relevant detailed categories. These include both common and more specific medical concepts (e.g., Tobacco Use, System Organ Site, Recreational Drug Use, Generic Name), designed to better reflect practical clinical information needs.
\item We conducted a methodologically consistent comparison of zero-shot, few-shot, and fine-tuned approaches using the same LLaMA3 backbone in granular MER.
\item For few-shot learning, we proposed two similarity-based example selection strategies: sentence-level and token-level embedding similarity, leveraging the domain-specific pre-trained BioBERT~\cite{lee_biobert_2020} model for embedding. 
\item We employed parameter efficient fine-tuning via LoRA~\cite{hu_lora_2021} to adapt LLaMA3 to the fine-grained MER task, and evaluated multiple LoRA configurations to assess the trade-off between training cost and model performance.
\item Our experiments showed that fine-tuning LLaMA3 achieved an F1 score of 81.24\% in granular MER, outperforming the best zero-shot and few-shot setups by 63.11\% and 35.63\%, respectively.
\end{itemize}


\section{Related Work}

\subsection{Medical Named Entity Recognition}
In MER, most methods treat the task as a sequence tagging problem. Early MER systems used BiLSTM~\cite{hochreiter_long_1997,schuster_bidirectional_1997} models paired with Conditional Random Fields (CRFs)~\cite{lafferty_conditional_1997} to predict entity labels across input sequences~\cite{huang_bidirectional_2015}. Over time, these models evolved with the introduction of transformer-based encoders like BERT, along with specialised variants such as BioBERT~\cite{lee_biobert_2020}, and PubMedBERT~\cite{gu_domain-specific_2022}, which are pre-trained on biomedical corpora to better capture domain-specific language. Subsequently, domain-specific pre-trained language models demonstrated state-of-the-art performance in MER tasks.

\subsection{LLM-based NER with In-Context Learning}
LLMs have shown impressive capabilities in performing various NLP tasks including entity recognition, often without the need for task-specific training data. LLMs support a novel learning paradigm known as in-context learning (ICL)~\cite{dong_survey_2024} where the model performs tasks by interpreting information provided within the input prompt without any changes to its internal parameters. Rather than relying on traditional training or fine-tuning, ICL allows the model to generate outputs based on task instructions or, when available, example demonstrations. The prompt serves as a guide, activating the model’s ability to reason or generalise based on the context it receives. 

Within ICL paradigm, two widely used settings are zero-shot and few-shot learning. In zero-shot learning, the model is given only a task description or natural language instruction, without any labelled examples or demonstrations. It relies entirely on its pre-trained knowledge and ability to interpret the instruction to generate a response. This setting tests the model's generalisation capability and its understanding of task intent without additional guidance. Few-shot learning refers to the setting where a pre-trained LLM is given a small set of input-output examples during inference to guide its performance, without any modification to the model’s weights~\cite{brown_language_2020}. This approach uses a small number of examples provided in the prompt as contextual guidance, allowing the model to make predictions with minimal prior training, which is an essential capability for real-world scenarios where annotated data is scarce. These approaches allow rapid adaptation to new tasks, with minimal supervision.

In the context of applying ICL to NER, prior work has explored both few-shot and zero-shot capabilities. A recent study~\cite{wang_gpt-ner_2023} proposed GPT-NER, a method that improves few-shot performance by selecting semantically similar examples for prompts using SimCSE~\cite{ gao_simcse_2022}, a contrastive sentence embedding model, instead of traditional encoders like RoBERTa-large~\cite{liu_roberta_2019}. This approach demonstrated strong results in low-resource conditions, achieving comparable performance to fully supervised BERT-based models on standard NER benchmark datasets of CoNLL-2003~\cite{ sang_introduction_2003} and OntoNotes5.0~\cite{pradhan_towards_2013}. A more recent study~\cite{tang_fsponer_2025} aimed to improve few-shot NER performance by using a TF-IDF-based example selection strategy. This method retrieves the most relevant in-context examples based on lexical similarity to the input sentence, thereby enhancing model accuracy. While the overall framework resembles GPT-NER, it replaces the sentence embedding component with a simpler TF-IDF approach. Their method was evaluated on three domain-specific NER datasets and outperformed GPT-NER in all cases. Separately, another study~\cite{xie_empirical_2023} focused on enhancing zero-shot NER by restructuring the task into simpler subtasks and applying reasoning strategies within ChatGPT, achieving improved performance across seven datasets, spanning both general and domain-specific NER tasks.

\section{Methodology}

\subsection{Task Definition}

We formulated a granular \textbf{MER} task on a corpus of clinical \textbf{discharge summaries}. Each document consists of free-text narratives describing a patient's hospital course and was divided into $n$ sentences. Given a sentence $S = (w_1, w_2, w_3, \ldots, w_n)$ composed of tokens, the model was required to predict a set of labeled entities $Y = \{ y_1, y_2, \ldots, y_k \}$, where each entity $y_j$ is defined as:
\[
y_j = (\text{span text}, \text{entity type},\text{start offset},\text{end offset})
\]
The objective of the task was to accurately extract all relevant medical entities from each sentence in the document, including their exact text spans, corresponding entity types, and the precise character positions (start and end offsets) where each entity occurs within the sentence.

\subsection{LLM for Medical Entity Recognition}

We employed the open-source LLaMA3~\cite{meta_llama_2024} model for fine-grained MER task, specifically the 8B instruction-tuned variant, which balances performance with inference and fine-tuning speed. LLaMA3 is an auto-regressive transformer-based model. The tuned version benefits from supervised fine-tuning (SFT) and reinforcement learning to optimise performance. The basic information of the model is presented in Table~\ref{table:llama3_8b}.

\begin{table}[tb]
\centering
\caption{The Basic Information of LLaMA3-8B-Instruct}
\vspace{8mm}
\begin{tabular}{|l|c|}
\hline
\textbf{Hyperparameter} & \textbf{Value} \\
\hline
Model Size (Parameters) & 8B \\
Layers & 32 \\
Model Dimension & 4,096 \\
FFN Dimension & 14,336 \\
Attention Heads & 32 \\
Key/Value Heads & 8 \\
Max Context Length & 8k \\
\hline
\textbf{Knowledge cutoff} & \textbf{March, 2023} \\
\hline
\end{tabular}
\label{table:llama3_8b}
\end{table}

\subsection{Dataset Preparation}

We used discharge summaries from the i2b2 dataset, which contained a total of 310 documents—190 for training (7,446 sentences, 91,110 words) and 120 for testing (5,665 sentences, 75,287 words). We used only the raw discharge summaries from the dataset, excluding the original i2b2 annotations. Instead, to capture more granular medical concepts relevant to downstream applications in real world clinical scenarios, we annotated 18 types of detailed medical entities: \textbf{System Organ Site, Alcohol Consumption, Allergies, Gender, Race/Ethnicity, Recreational Drug Use, Tobacco Use, DX Name, Brand Name, Generic Name, Procedure Name, Test Name, Treatment Name, Time to DX Name, Time to Medication Name, Time to Procedure Name, Time to Test Name, and Time to Treatment Name} (please see Appendix for detailed definitions of each entity type).

\begin{figure}[tb]
\centering
\includegraphics[width=\linewidth]{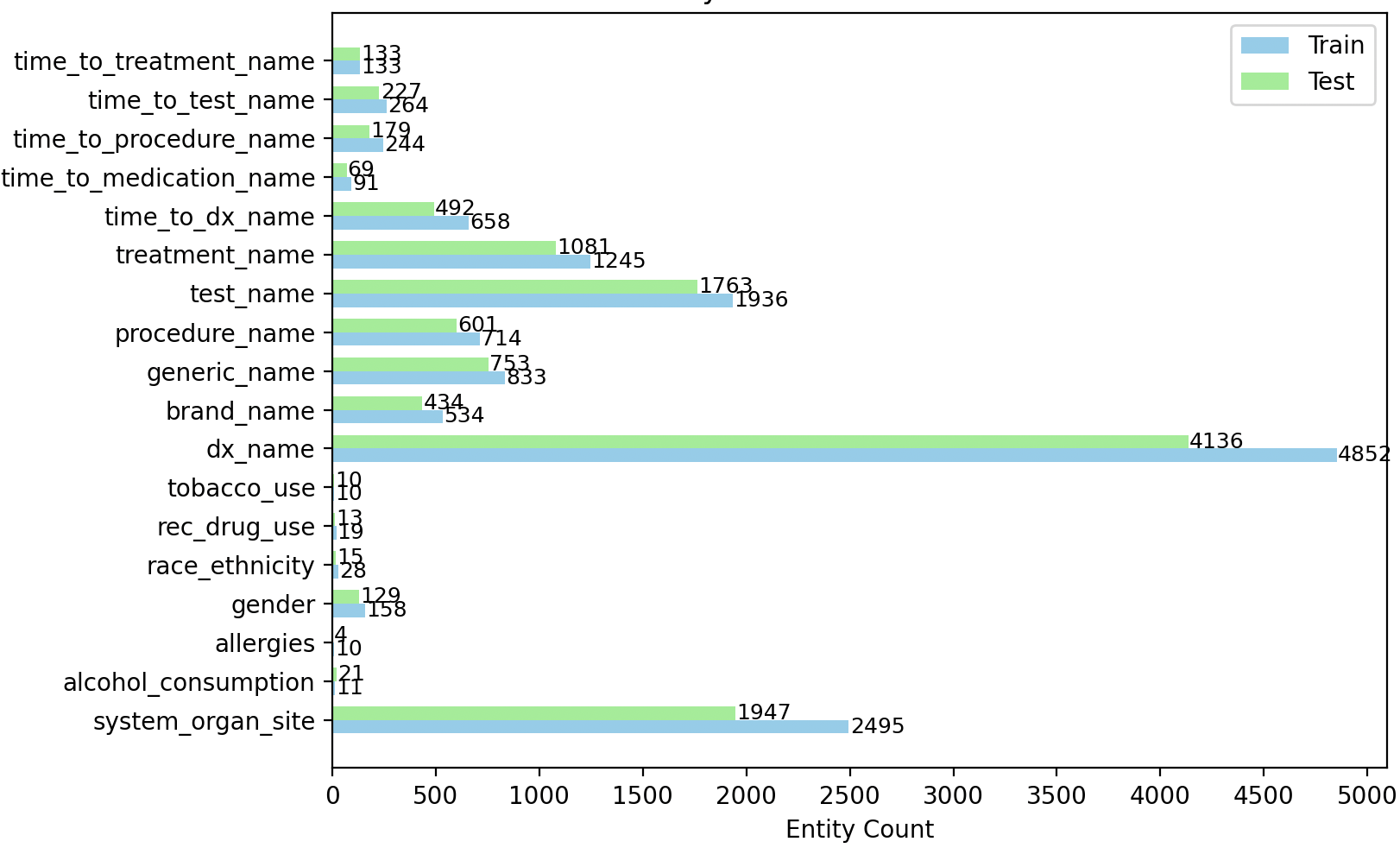}
\caption{Entity Breakdown: Train (14,235 entities) vs. Test (12,007 entities)}
\label{fig:ent_breakdown}
\vspace{8mm}
\end{figure}

Our entity schema was inspired by the level of detail provided by Amazon Comprehend Medical (ACM)\footnote{\url{https://https://aws.amazon.com/comprehend/medical/}}. To aid the annotation process, we first pre-annotated the discharge summaries using the ACM. These initial annotations were then reviewed and refined by our medical professionals experienced with unstructured clinical notes, using an open-source tool, Doccano~\cite{nakayama_doccano_2018}, which provided a user-friendly interface for manual annotation.
The statistics of entities in the annotated dataset are shown in Figure~\ref{fig:ent_breakdown}.

\subsection{Few-shot Example Selection Methods}
We explored strategies for selecting few-shot examples that were semantically aligned with the input sentence to enhance the model's performance in MER. This was particularly critical in few-shot learning, where our baseline model, LLaMA3, may have faced challenges with domain-specific medical entities because it primarily relied on general-purpose knowledge acquired during pre-training. To address this, we adopted similarity-based retrieval techniques, selecting examples based on embedding-level similarity to the input sentence, measured using cosine similarity. Specifically, we incorporated two approaches: sentence-level embedding similarity and token-level embedding similarity. For both approaches, we leveraged domain-specific embeddings generated by a pre-trained BioBERT model to better capture the characteristics and complexity of clinical language. 

\subsubsection{Top-$k$ Sentence-level embedding similarity}

In this approach, we aimed to select the top-$k$ most similar sentences from the training dataset based on sentence-level embeddings. Each sentence in the dataset and the input sentence were represented by their corresponding embeddings, denoted as $\mathbf{e}_t$ for sentence $t$ from the training set and $\mathbf{e}_{\text{input}}$ for the input sentence. We calculated the cosine similarity between the input sentence embedding $\mathbf{e}_{\text{input}}$ and the embeddings of all other sentences $\mathbf{e}_t$, for $t \in \{1, 2, \ldots, T\}$, where $T$ was the total number of sentences in the training dataset. The cosine similarity was defined as:
\[
\text{sim}(\mathbf{e}_t, \mathbf{e}_{\text{input}}) = \frac{\mathbf{e}_t \cdot \mathbf{e}_{\text{input}}}{\|\mathbf{e}_t\| \|\mathbf{e}_{\text{input}}\|}
\]
Where $\mathbf{e}_t \cdot \mathbf{e}_{\text{input}}$ was the dot product between the embeddings of sentence $t$ from the training set and the input sentence.
$\|\mathbf{e}_t\|$ and $\|\mathbf{e}_{\text{input}}\|$ were the Euclidean norms of the sentence embeddings.
Once the similarity scores were computed for all sentences, we ranked them in descending order of similarity and selected the top-$k$ sentences with the highest similarity scores:
\[
\text{Top-}k = \arg\max_{\{t_1, t_2, \dots, t_k\}} \left( \text{sim}(\mathbf{e}_t, \mathbf{e}_{\text{input}}) \right)
\]
These top-$k$ sentences were then used as the few-shot examples for the model.

\subsubsection{Top-$k$ token-level embedding similarity}

In this method, we identified the top-$k$ most similar token sequences to the input sentence by comparing token-level embeddings. Each token in a sentence was represented by its embedding $\mathbf{t}_{t,j}$ for token $j$ in sentence $t$, and similarly, the input sentence tokens were represented by their embeddings $\mathbf{t}_{\text{input},j}$. For each sentence in the training dataset, we calculated the cosine similarity between the token embeddings of the input sentence and the corresponding token embeddings in the sentence being evaluated. We defined the cosine similarity between token $j$ in sentence $t$ and token $j$ in the input sentence as:
\[
\text{sim}(\mathbf{t}_{t,j}, \mathbf{t}_{\text{input},j}) = \frac{\mathbf{t}_{t,j} \cdot \mathbf{t}_{\text{input},j}}{\|\mathbf{t}_{t,j}\| \|\mathbf{t}_{\text{input},j}\|}
\]
Where $\mathbf{t}_{t,j} \cdot \mathbf{t}_{\text{input},j}$ was the dot product between the embeddings of token $j$ in sentence $t$ from the training set and the token $j$ in the input sentence.
 $\|\mathbf{t}_{t,j}\|$ and $\|\mathbf{t}_{\text{input},j}\|$ were the Euclidean norms of the token embeddings.
To calculate the overall similarity for the entire sentence, we aggregated the token-level similarities by averaging them:
\[
\text{sim}_{\text{sentence}}(\mathbf{e}_t, \mathbf{e}_{\text{input}}) = \frac{1}{n} \sum_{j=1}^{n} \text{sim}(\mathbf{t}_{t,j}, \mathbf{t}_{\text{input},j})
\]
Where $n$ was the total number of tokens in sentence $t$. After calculating the sentence-level similarity for all sentences in the training dataset, we selected the top-$k$ sentences by ranking them based on their similarity scores:
\[
\text{Top-}k = \arg\max_{\{t_1, t_2, \dots, t_k\}} \left( \text{sim}_{\text{sentence}}(\mathbf{e}_t, \mathbf{e}_{\text{input}}) \right)
\]
These top-$k$ sentences were then chosen as the few-shot examples for the model.

\subsection{Prompt Structure}

We standardised prompt structures across different learning settings to ensure fair and controlled comparisons. Variations in prompt design can significantly influence model behaviour and final accuracy, potentially leading to inconsistent evaluations across different learning settings.

\subsubsection{Baseline prompt}
\begin{figure}[tb]
\centering
\includegraphics[width=\linewidth]{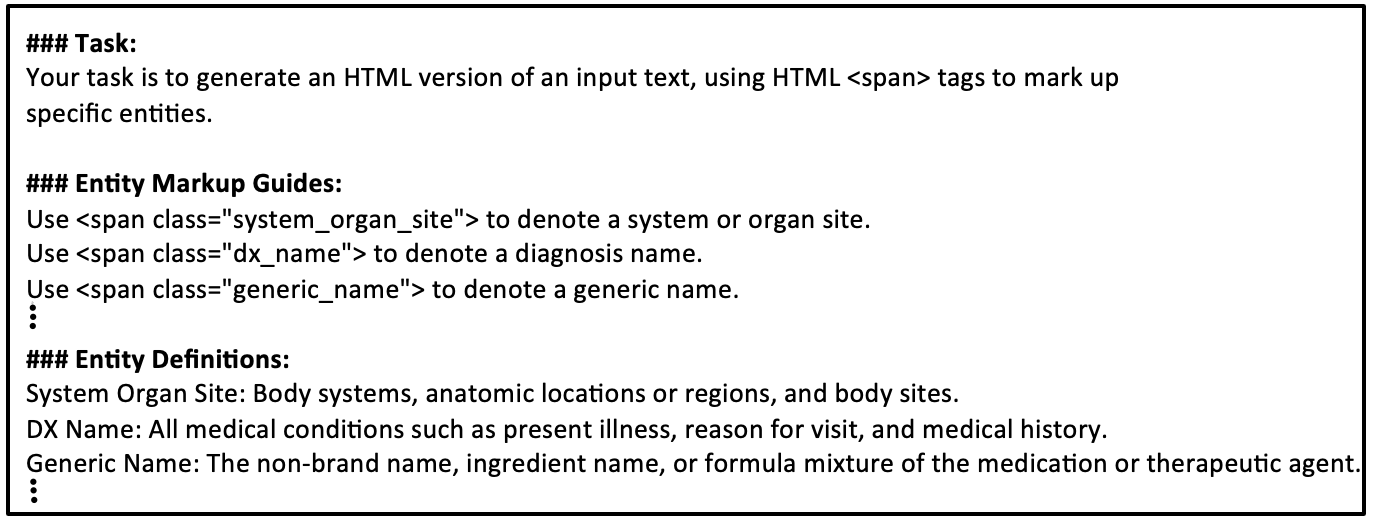}
\caption{Baseline Prompt Structure}
\label{fig:b_prompt}
\vspace{8mm}
\end{figure}

The baseline prompt, applied across fine-tuning, zero-shot, and few-shot settings, was composed of three main components: Task Description, Entity Markup Guidelines, and Entity Definitions. This prompt style was inspired by previous studies~\cite{hu_improving_2024,hu_information_2025}. The detailed structure of the baseline prompt is illustrated in Figure~\ref{fig:b_prompt}.

\subsubsection{Baseline prompt + strict entity extraction guidelines}
\begin{figure}[tb]
\centering
\includegraphics[width=\linewidth]{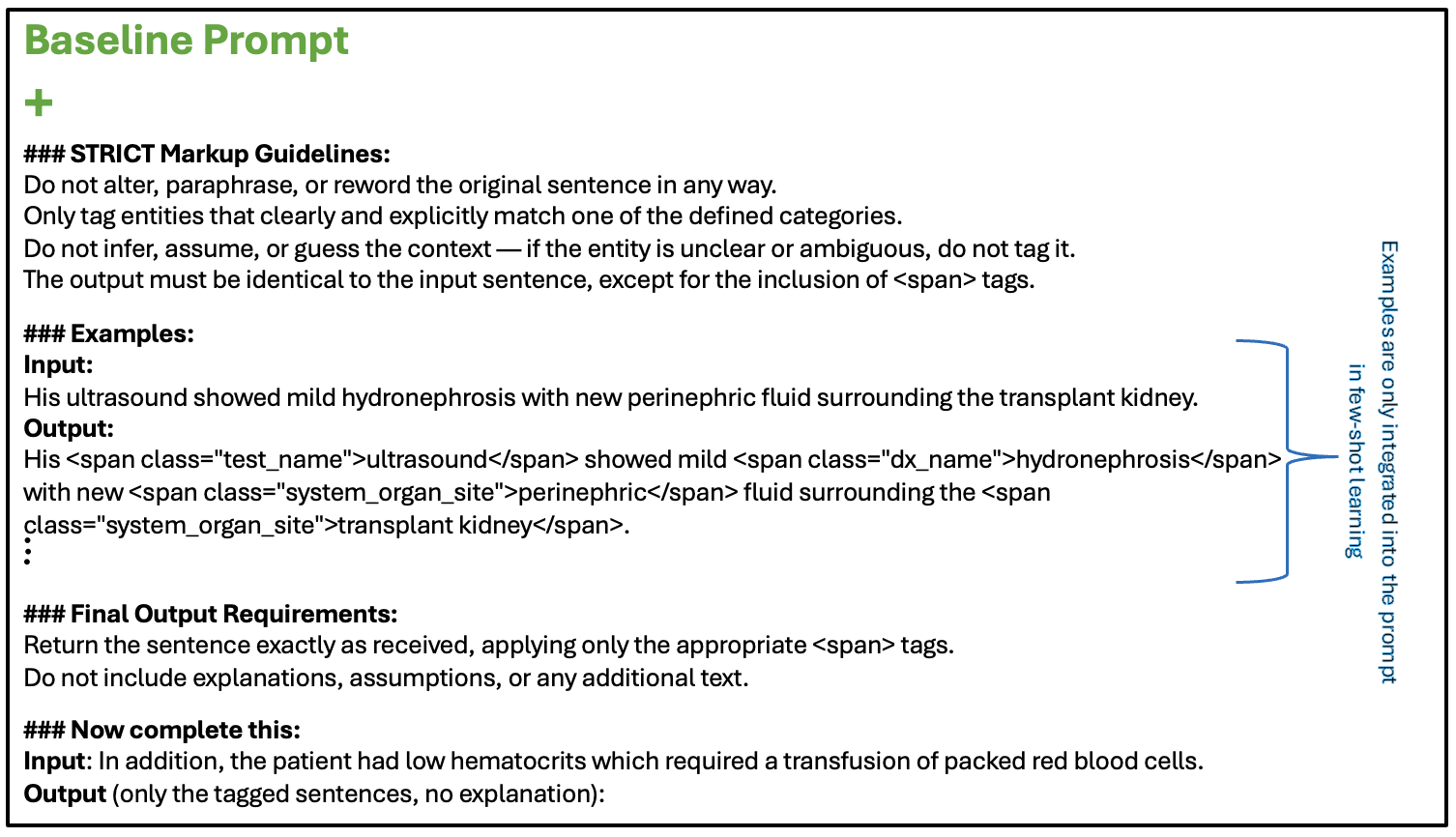}
\caption{Enhanced Prompt Structure Used in Zero/Few-shot Learning}
\label{fig:e_prompt}
\vspace{8mm}
\end{figure}

To further standardise final outputs, particularly in zero-shot and few-shot settings, we extended the baseline prompt by adding strict formatting guidelines. These guidelines enforced a structured output format, instructing the model to avoid including reasoning, explanations, or assumptions in the final output, and not to alter the given input sentence. This is critical for ensuring that entity extraction remains strictly grounded in the input. The complete structure of this enhanced prompt is illustrated in Figure~\ref{fig:e_prompt}. In few-shot learning, the \textbf{Examples block} were included; in zero-shot learning, no examples were provided, allowing us to evaluate the model's performance based solely on the prompt instructions.

\subsection{Fine-Tuning}
Previous studies~\cite{hu_improving_2024,wang_gpt-ner_2023, tang_fsponer_2025} have demonstrated the effectiveness of few-shot learning for NER in both general and domain specific scenarios, often using proprietary models like the GPT series. However, for domain-specific NER, particularly in the granular MER task, there is a lack of direct comparison between few-shot, zero-shot, and fine-tuning approaches on open-source LLMs. To address this gap, we fine-tuned the baseline LLaMA3 8B Instruct model to enable direct comparison among the three learning paradigms. The training was conducted on a training dataset of 7,446 sentences. We then evaluated the performance of the baseline model in both zero-shot and few-shot learning settings, and the performance of the fine-tuned model during inference on a test set comprising 5,665 sentences, to assess their effectiveness in granular MER.

\subsubsection{Preprocessing for training dataset}
\begin{figure}[tb]
\centering
\includegraphics[width=\linewidth]{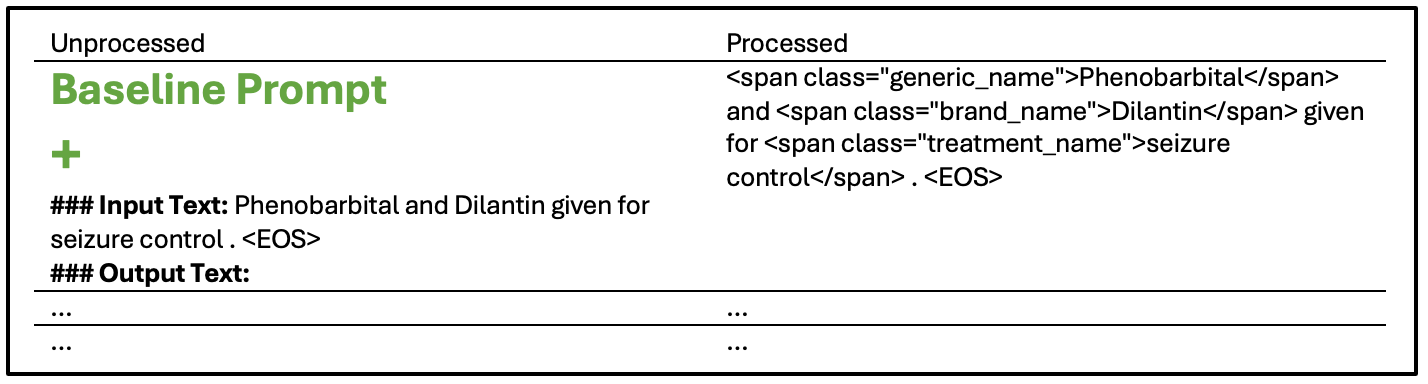}
\caption{Training Dataset Used in Model Fine-Tuning }
\label{fig:train_set}
\vspace{8mm}
\end{figure}

The training data consisted of two columns: Unprocessed and Processed. The Unprocessed column combined the baseline prompt with the input sentence, while the Processed column contained the ground truth—i.e., the input sentence annotated with the corresponding entity types, as illustrated in Figure~\ref{fig:train_set}. During inference, we passed the same prompt described in the Unprocessed column along with a new input sentence from the test set, and the fine-tuned model generated the output sentence with the predicted corresponding entity types.

\subsubsection{Low-rank adaptation}
For parameter-efficient fine-tuning, we employed LoRA~\cite{hu_lora_2021} which enhanced fine-tuning efficiency by expressing weight updates using two compact matrices derived through low-rank decomposition. These smaller matrices were trained to adapt the model to new data, allowing minimal overall parameter changes. Meanwhile, the original weight matrix stayed frozen and was not modified during training. To generate the final output, the original and adapted weights were combined. During inference, this merged representation was used to produce the model’s response for new, unseen input sentences. This approach significantly reduces memory consumption and computational cost, making it suited for fine-tuning large language models like LLaMA3 on domain-specific small dataset. In our implementation, we used the following different LoRA configurations: a LoRA attention dimension (r) of 8, 16, 32 and 64, the alpha parameter for LoRA scaling (lora\_alpha) of 64 and 128, and the dropout probability for LoRA layers (lora\_dropout) of 0.05. We applied LoRA to all 7 modules of the baseline model (q\_proj, k\_proj, v\_proj, o\_proj, down\_proj, up\_proj, gate\_proj), and also applied it to only 4 attention modules (q\_proj, k\_proj, v\_proj, o\_proj) of the baseline model. The bias parameter was set to 'none' to avoid updating biases during training. The task type was specified as 'CAUSAL\_LM' to align with the autoregressive nature of the baseline model. We then observed the number of trainable paraameters, training time taken, and the accuracy of the fine-tuned models with different LoRA configurations; these are discussed in the Results and Discussion section.

\subsection{Evaluation Matrix}
For evaluation, we employed precision, recall, and F1-score metrics to assess entity extraction performance using exact matching. A predicted entity was considered a true positive if its span text and entity type matched the ground truth, and the start and end offsets of the span were within ±2 characters of the ground truth span. The ±2 character tolerance was introduced to minimise sensitivity to minor discrepancies such as extra spaces in the output, which could otherwise cause mismatches when comparing the start and end offsets of predicted spans with the ground truth. All unmatched predictions were counted as false positives, and unmatched ground truth entities as false negatives. These counts were used to compute overall precision, recall, and F1-score, as well as per-entity-type scores. 
Let a predicted entity be represented as \( p = (t_p, s_p, e_p, l_p) \) and a ground truth entity as \( g = (t_g, s_g, e_g, l_g) \),  
where \( t \) is the span text, \( s \) is the start offset, \( e \) is the end offset, and \( l \) is the entity label (type).
A predicted entity \( p \) is considered a \textbf{True Positive (TP)} if:
\[
t_p = t_g, \quad l_p = l_g, \quad |s_p - s_g| \leq 2, \quad |e_p - e_g| \leq 2
\]

Additionally, we tracked the Invalid Entity Count (\%), defined as the proportion of predicted entities that fall outside our set of 18 predefined medical entity types to observe the LLM's hallucination and its ability to follow prompt guidelines in zero-shot learning, few-shot learning, and fine-tuned model settings.

\section{Experiments}

We detailed the experimental setup, results, and analysis.

\subsection{Experimental Setting}

Fine-tuning of LLaMA3 8B Instruct was performed on our institution's HPC, using single-node training without distributed training. Each node was equipped with an NVIDIA A30 GPU, 8 CPU cores, and 32 GB of both CPU and GPU memory. We recorded the training time for each combination of LoRA configuration to analyse performance and efficiency.
The environment was configured with PyTorch and Hugging Face Transformers libraries. This setup ensured consistent and reproducible performance for evaluating fine-tuning strategies with different LoRA configurations. We further applied 4-bit quantisation to load the base model’s weights in a 4-bit quantised format with bfloat16 compute precision using the BitsAndBytes library, reducing memory usage while maintaining high model performance during fine-tuning.

 All fine-tuning experiments were conducted with 2 epochs. Preliminary runs with 5 epochs revealed that the training loss plateaued after epoch 2, showing no significant decrease. To optimise training time and computational resources while ensuring fair evaluation across different LoRA configurations, we consistently limited fine-tuning to 2 epochs for all experiments.

To perform zero-shot and few-shot evaluation, we used a locally deployed baseline LLaMA3 8B Instruct model via Ollama. The local machine’s specifications were: 
Apple M1 Pro with an 8-core CPU, 14-core GPU, 16-core Neural Engine, and 16 GB of unified memory.

\subsection{Results and Discussions}

\begin{table}[tb]
\caption{Comparison of LoRA configurations: trainable parameters, training time, and F1 score}
\vspace{5mm}
\centering
\renewcommand{\arraystretch}{1.1}
\begin{tabular}{p{2.0cm} @{\hspace{3mm}} r @{\hspace{3mm}} r @{\hspace{3mm}} r @{\hspace{3mm}} r @{\hspace{3mm}} p{1.2cm} @{\hspace{3mm}} r}
\toprule
Modules & $r$ & $\alpha$ & \shortstack{Trainable\\Parameters} & \shortstack{Time\\(2 epoches)} & F1 \\
\midrule
q,k,v,o         & 8  & 64   & 6.81M    &  2h 59m & 78.57 \\
q,k,v,o         & 16 & 64   & 13.63M   &  2h 59m & 78.75 \\
q,k,v,o         & 32 & 64   & 27.26M   &  2h 59m & 78.61 \\
q,k,v,o         & 64 & 128  & 54.52M   &  2h 59m & 79.41 \\
all (7 modules) & 8  & 64   & 20.97M   &  3h 33m & 81.11 \\
all (7 modules) & 16 & 64   & 41.94M   &  3h 36m & 81.12 \\
all (7 modules) & 32 & 64   & 83.88M   &  3h 41m & 80.82 \\
all (7 modules) & 64 & 128  & 167.77M  &  3h 47m & \textbf{81.24}  \\
\bottomrule
\end{tabular}
\label{table:LoRA}
\end{table}

\begin{figure*}[tb]
\centering
\includegraphics[width=\linewidth]{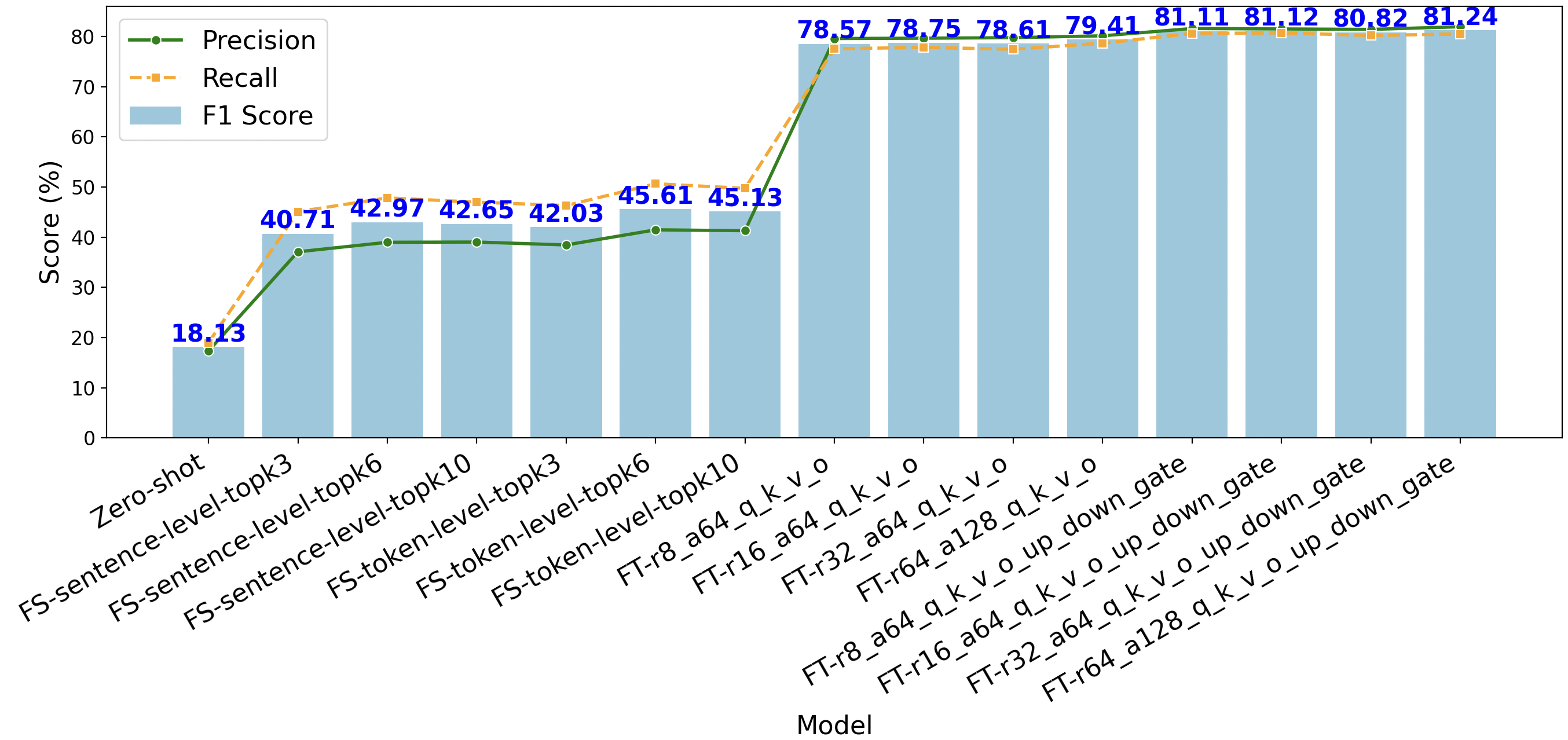}
\caption{Overall F1-Score, Precision, and Recall Across Different Models. The numbers indicate value of F1-scores. Model names beginning with FS represent few-shot learning approaches, with suffixes indicating the example selection method (sentence-level or token-level embedding similarity) and topk denoting the number of examples used. Models prefixed with FT refer to fine-tuned models, annotated with their respective LoRA configurations: r denotes the LoRA rank dimension, a indicates the scaling factor (lora\_alpha), and the remaining components specify the modules where LoRA was applied}
\label{fig:overall_f1}
\vspace{5mm}
\end{figure*}

The performance comparison 
(Figure~\ref{fig:overall_f1} and Table~\ref{table:LoRA})  
reveals a clear distinction between the three learning paradigms we were investigating: zero-shot, few-shot (FS) and 
fine-tuned (FT)  configurations. Fine-tuned models consistently outperformed zero- and few-shot baselines across all evaluation metrics --- F1 score, precision, and recall. Overall, fine-tuned models achieved F1 scores in the range of approximately $80\% \pm 1.2$, compared to $43.5\% \pm 1.7$ for few-shot baselines and 18.13\% for the zero-shot baseline, highlighting the substantial performance gains from task-specific fine-tuning. 

Among the fine-tuned models, the best-performing model (configured with r = 64, lora\_alpha = 128, and modules = [q, k, v, o, down, up, gate]) achieved the highest overall F1 score of 81.24\%, indicating a strong balance between precision (81.96) and recall (80.52). Our results showed that applying LoRA adapters into all 7 modules of the baseline model consistently outperformed configurations where adapters were applied only to the 4 attention modules, as shown in Table~\ref{table:LoRA}. Additionally, increasing the LoRA dimension size (r) did not yield clinically significant performance gains in F1 score. 
These findings suggest that adapter placement has a greater impact on performance than simply increasing the size of r. 

\begin{table}[tb]
\caption{Wilcoxon Signed-Rank Test Results for Model Comparisons Between Group 1 (Smaller r) and Group 2 (Larger r).\\
Models in Pairs 1 to 3 were fine-tuned by applying LoRA adapters to the 4 attention modules (q, k, v, o). \\Models in Pairs 4 to 6 were fine-tuned by applying LoRA adapters to all 7 modules (q, k, v, o,up, down, gate).} 
\vspace{5mm}
\centering
\renewcommand{\arraystretch}{1.1}
\begin{tabular}{p{1.3cm} @{\hspace{3mm}} r @{\hspace{3mm}} r @{\hspace{3mm}} r @{\hspace{3mm}} r @{\hspace{3mm}} p{1.2cm} @{\hspace{3mm}} r}
\toprule
Pair & \shortstack{Group 1\\(smaller r)} & \shortstack{Group 2\\(larger r)} & \shortstack{Wilcoxon W} & \shortstack{p-value} & \shortstack{Statistically\\Significant} \\
\midrule
Pair 1  & r=8 & r=16 & 55.0    &  0.5014 & No \\
Pair 2  & r=8 & r=32  & 84.0   &   0.9661 & No \\
Pair 3  & r=16 & r=32   & 78.0   &  0.7660 & No \\
Pair 3  & r=8 & r=16  & 67.0   &  0.9587 & No \\
Pair 5  & r=8 & r=32   & 66.0   &  0.9176 & No \\
Pair 6  & r=16 & r=32   & 66.0   &  0.6191 & No \\
\bottomrule
\end{tabular}
\label{table:LoRA}
\end{table}

Among the few-shot approaches, the configuration that integrated few-shot examples selected via token-level similarity with top-k = 6 yielded the best result, achieving an F1 score of 45.61\%. This highlights the advantage of token-level over sentence-level embedding similarity for sample selection. Since MER is a token-level task, token-level similarity is more effective for choosing closely related examples to support few-shot learning. Additionally, we examined the impact of increasing the number of few-shot examples during inference, starting with 3 and scaling up to 6 and then 10 (see Figure~\ref{fig:overall_f1} FS models with different \texttt{topk} values), in both sentence-level and token-level similarity selection methods. In both methods, increasing from 3 to 6 examples led to a 2–3\% improvement in F1 score, while increasing from 6 to 10 examples did not result in further gains and in some cases slightly reduced performance compared to the 6-sample configuration. 

In contrast, the zero-shot baseline performed substantially worse, with an F1 score of just 18.13\%, clearly illustrated in Figure~\ref{fig:overall_f1}, underscoring the effectiveness of model adaptation strategies --- particularly fine-tuning --- for enhancing domain-specific downstream performance in granular MER tasks. Furthermore, an analysis of precision and recall across different training paradigms revealed a consistent pattern. In all zero-shot and few-shot settings, recall was consistently higher than precision, and the model tended to over-generate or over-predict entities. In all fine-tuned settings, precision consistently exceeded recall, as the model learned task-specific constraints more effectively, thereby reducing over-generation. 

\begin{figure*}[tb]
\centering
\includegraphics[width=\linewidth]{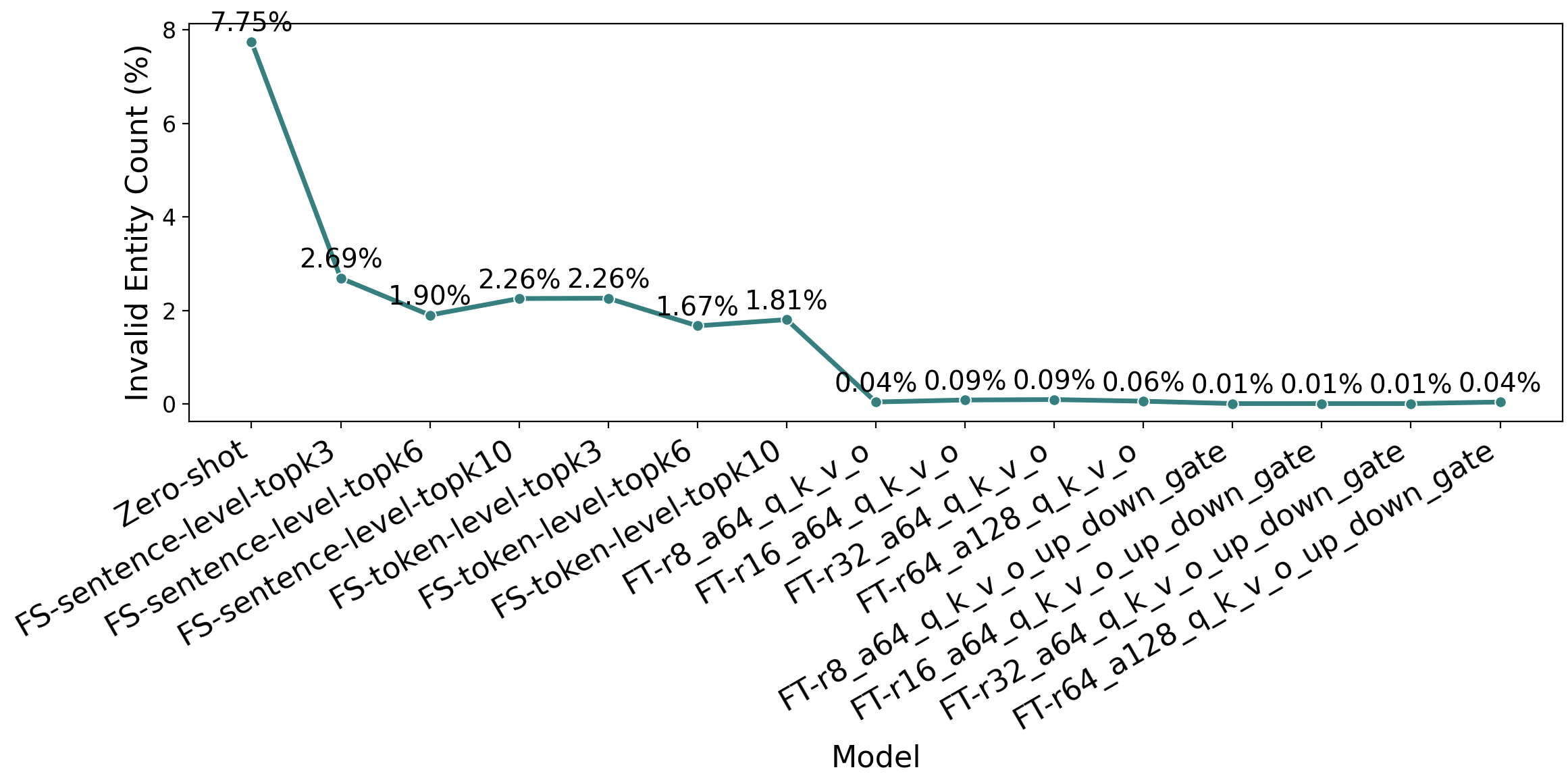}
\caption{Invalid Entity Count (in \% of Total Predicted Entities) Across Different Models}
\label{fig:invalid_ent}
\vspace{5mm}
\end{figure*}

In addition to standard performance metrics such as F1 score, we assessed the quality of entity predictions by measuring the Invalid Entity Count (\%), defined as the proportion of extracted entities that did not conform to valid 18 entity types predefined in the prompt guideline (Figure~\ref{fig:invalid_ent}). 
The results showed that zero-shot inference exhibited the highest rate of invalid entities at 7.75\%, reflecting the model's limited capacity for domain adaptation without fine-tuning or tailored prompting. This higher rate also suggests a greater risk of hallucination --- producing clinically irrelevant or unsupported outputs --- under zero-shot settings. Few-shot configurations substantially reduced the invalid entity rate, particularly when leveraging token-level similarity. The few-shot setup that integrated examples selected via token-level similarity with top-k = 6 achieved the lowest invalid rate among few-shot methods at 1.67\%, outperforming the sentence-level counterpart (1.90\% for top-k = 6). However, all fine-tuned models surpassed few-shot approaches by a significant margin. Notably, the fine-tuned models using the LoRA adapters in all 7 modules (q, k, v, o, down, up, gate) yielded the lowest invalid entity rate ranging from 0.01\% to 0.04\%, indicating highly reliable entity extraction. Other fine-tuned models injecting LoRA adapters into only 4 attention modules (q,k,v,o) also maintained comparably low rates, ranging from 0.04\% to 0.09\%. These findings reinforce the superiority of fine-tuning not only in overall performance but also in preserving the structural integrity of predicted outputs.

\begin{figure*}
\centering
\includegraphics[width=\linewidth]{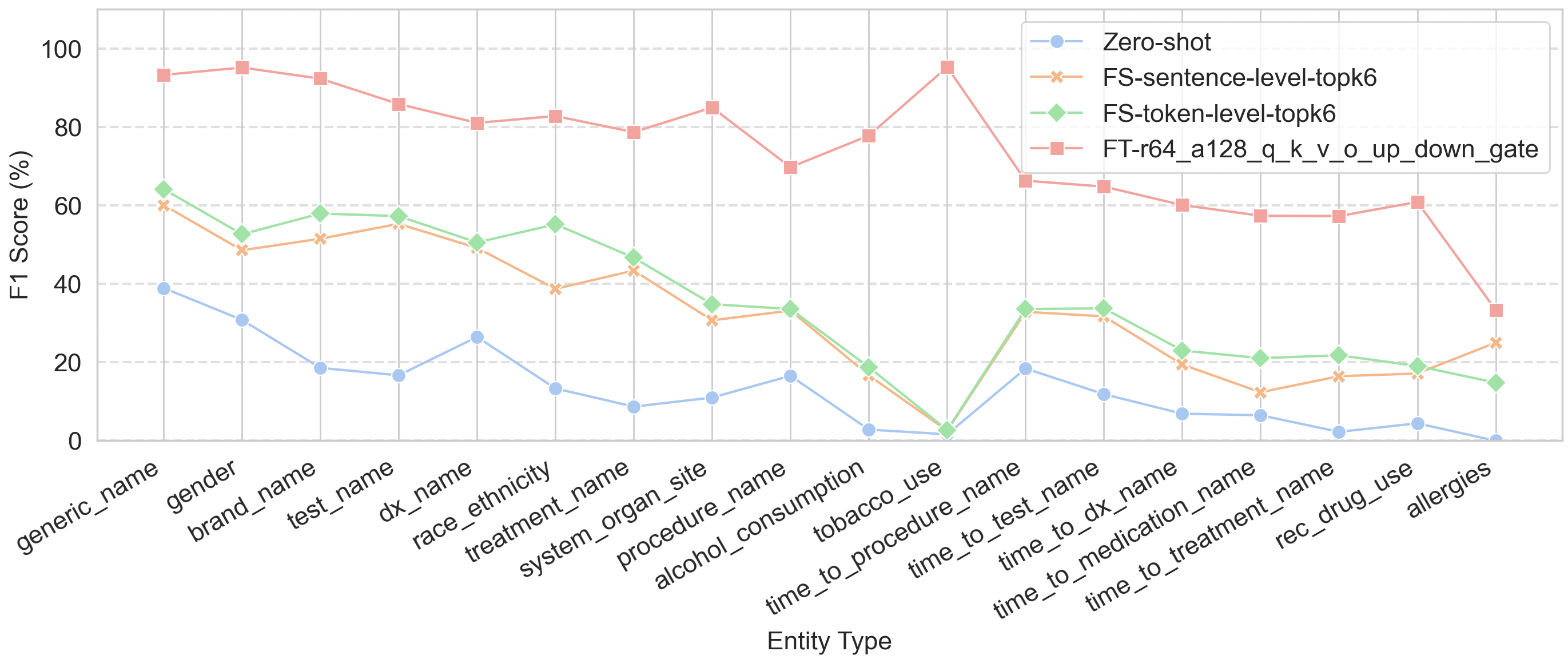}
\caption{Per-Entity F1 Score (Selected Models)}
\label{fig:ent_f1}
\vspace{8mm}
\end{figure*}

To further investigate model performance at the level of individual entities (Figure~\ref{fig:ent_f1}), we conducted a per-entity F1 score analysis across selected models, demonstrating clear performance gaps between zero-shot, few-shot, and fine-tuned configurations. The zero-shot setup struggled across most entity types. In contrast, few-shot models, especially those leveraging token-level similarity (top-k = 6), showed marked improvements across nearly all entity types. Notably, entities such as generic\_name (F1 = 64.1\%), test\_name (F1 = 57.3\%), and dx\_name (F1 = 50.5\%), reflecting the advantage of token-level granularity in example selection. However, the fine-tuned model (LoRA r=64, lora\_alpha=128, modules = [q, k, v, o, down, up, gate]) consistently outperformed both few-shot and zero-shot methods by a wide margin in all entity types. It achieved F1 scores exceeding 84.9\% for system\_organ\_site, 93.2\% for generic\_name, and 95.2\% for tobacco\_use, indicating highly effective adaptation to the task's entity-specific nuances. Even for lower-frequency entities such as alcohol\_consumption and rec\_drug\_use, the fine-tuned model maintained strong F1 scores (77.7\% and 60.8\%, respectively). These results highlight that while few-shot learning offers notable improvements over zero-shot inference, fine-tuning remains essential for achieving high-precision entity extraction across both frequent and rare categories.

\section{Conclusion and Future Work}

In this study, we conducted a comprehensive evaluation of the open-source large language model LLaMA3 8B, balancing inference speed, training efficiency, and accuracy for the fine-grained MER task. We assessed the model’s performance across three popular learning paradigms --- zero-shot, few-shot, and fine-tuning --- using a consistent prompt structure and the same backbone model to ensure fair comparison. To support continued research in MER, we developed a new fine-grained MER dataset, aligned with real clinical downstream needs, and serving as a benchmark for future work. Our experiments showed that, in few-shot settings, selecting examples based on token-level embedding similarity yielded better accuracy than sentence-level similarity for granular NER tasks. Among all settings, fine-tuning achieved the best performance, reaching an F1 score of 81.24\%, almost double that of few-shot learning, and quadruple the score of zero-shot learning. Notably, we found that integrating LoRA adapters into all linear modules of the baseline model rather than just attention layers led to improved performance.

This work addresses a methodological gap by providing a unified evaluation of learning paradigms on a shared open-source LLM architecture for the fine-grained MER task, and contributes a new benchmark dataset to support further research. The paper source code including the dataset is on GitHub and will be available upon paper acceptance.

For future work, we plan to focus on improving MER performance, with the dual goal of increasing accuracy and reducing hallucination. We also intend to evaluate the generalisability of our best-performing models on real clinical data from our local health district. 
Furthermore, we aim to extend our research into relation extraction, including temporal relation modelling among identified medical entities, to enable richer representations of clinical concepts.






\bibliography{mybibfile}

\end{document}